\pdfoutput=1

\documentclass[11pt]{article}

\usepackage{tabularx}
\usepackage{makecell} 
\usepackage{booktabs}
\usepackage{todonotes}
\usepackage{tikz}

\usepackage{colortbl} 
\usepackage[preprint]{acl}

\usepackage{times}
\usepackage{latexsym}

\usepackage[T1]{fontenc}

\usepackage[utf8]{inputenc}

\usepackage{microtype}

\usepackage{inconsolata}

\usepackage{graphicx}

\usepackage{amsmath}
\usepackage{subcaption}
\title{AntLM: Bridging Causal and Masked Language Models}

\author{
 Xinru Yu\textsuperscript{\rm 1}\thanks{\ \ Equal contribution.},
 Bin Guo\textsuperscript{\rm 1}\footnotemark[1],
 Shiwei Luo\textsuperscript{\rm 3}\footnotemark[1],
 Jie Wang\textsuperscript{\rm 1},
 \textbf{
 Tao Ji\textsuperscript{\rm 2}\thanks{\ \ Corresponding authors.},
 Yuanbin Wu\textsuperscript{\rm 1}}\footnotemark[2]
\\
 \textsuperscript{1} School of Computer Science and Technology, East China Normal University\\
 \textsuperscript{2} School of Computer Science, Fudan University\\
 \textsuperscript{3} School of Computer Science and Technology, Harbin Engineering University
\\
 \small{
    {
    \href{mailto:xryu@stu.ecnu.edu.cn, binguo@stu.ecnu.edu.cn, jiewang@stu.ecnu.edu.cn, ybwu@cs.ecnu.edu.cn}{\{xryu@stu,binguo@stu,jiewang@stu,ybwu@cs\}.ecnu.edu.cn},
    \href{mailto:taoji@fudan.edu.cn}{taoji@fudan.edu.cn},
    \href{mailto:shiweiluomo@gmail.com}{shiweiluomo@gmail.com}
   }
   }
}

\begin{document}
\maketitle
\begin{abstract}
Causal Language Modeling (CLM) and Masked Language Modeling (MLM) are two mainstream learning paradigms based on Transformer networks, specifically the Decoder-only and Encoder-only architectures. 
The strengths of each paradigm in downstream tasks have shown a mix of advantages and disadvantages. 
In the past BabyLM Challenge 2023, although the MLM paradigm achieved the best average performance, the CLM paradigm demonstrated significantly faster convergence rates.
For the BabyLM Challenge 2024, we propose a novel language modeling paradigm named \textbf{AntLM}, which integrates both CLM and MLM to leverage the advantages of these two classic paradigms.
We chose the strict-small track and conducted experiments on two foundation models: BabyLlama, representing CLM, and LTG-BERT, representing MLM.
During the training process for specific foundation models, we alternate between applying CLM or MLM training objectives and causal or bidirectional attention masks.
Experimental results show that combining the two pretraining objectives leverages their strengths, enhancing overall training performance. 
Under the same epochs, AntLM$_\text{BabyLlama}$ improves Macro-average by 1\%, and AntLM$_\text{LTG-BERT}$ achieves a 2.2\% increase over the baselines.
\end{abstract}

\section{Introduction}
\label{sec:intro}

Language Modeling (LM) is a core task in NLP and a key technology for natural language understanding and generation, supporting a wide range of applications including machine translation~\cite{hendy2023good}, speech recognition~\cite{prabhavalkar2023end}, sentiment analysis~\cite{tan2023survey}, and information extraction~\cite{wei2023zero}.
Over the past decades, LM has seen significant development, evolving from simple models like n-grams~\cite{ngram} to more sophisticated models, such as recurrent neural networks~\cite{rnn}, long short-term memory networks~\cite{lstm}, and more recently, Transformer-based large language models (LLMs) like GPT~\cite{gpt2} and BERT~\cite{devlin2018bert}. 
LLMs have demonstrated human-like or even superhuman performance in language modeling.

However, the tremendous success of LLMs relies on learning from massive corpora, which is not as data-efficient and low-energy as human language learning.
The BabyLM Challenge 2023~\cite{warstadt2023papersbabylmchallenge} and 2024~\cite{choshen2024callpapers2ndbabylm} is a shared task over two consecutive years. 
It aims to encourage the discovery of more effective methods for training models using limited data.
Considering that a 13-year-old child has encountered fewer than 100 million words in their lifetime, the shared task has introduced the \textit{strict-small track}\footnote{Due to limitations in computational resources, we have not yet explored the \textit{strict track} and the \textit{multimodal track}.}.
These tracks confine pre-training data to 10 million and 100 million words. 
These datasets consist of child-accessible materials, such as books, conversations, and Wikipedia entries, to enhance the relevance of language model pre-training to human language learning processes. Compared to 2023, the 2024 competition removed the Children’s Book Test~\cite{Childrens_Book_Test} and QCRI Educational Domain Corpus datasets~\cite{QED}. The 2024 competition also reduced the proportion of OpenSubtitles~\cite{OpenSubtitles} dataset while increasing the proportions of CHILDES~\cite{CHILDES} and Project Gutenberg~\cite{Gutenberg} datasets.

The current investigation of LMs primarily adopts two predominant modeling paradigms: Causal Language Models (CLMs), represented by GPT~\cite{gpt2}, and Masked Language Models (MLMs), represented by BERT~\cite{devlin2018bert}. CLMs employ next-token prediction as their training objective, which is predicting the next token given the preceding context, and they perform exceptionally well on generative tasks. On the other hand, the training objective of MLM is the random selection and masking of some tokens in the input text, following which the model is trained to predict the original unmasked tokens. Due to its global information modeling capabilities, this approach excels in tasks necessitating the capture of bidirectional contextual information, such as text classification. Considering these modeling paradigms' strengths, this paper raises an important question: Could the two modeling methodologies be seamlessly integrated?

Intuitively, performing the MLM task allows the model to learn bidirectional contextual encoding of text, while the CLM task enables the model to predict and generate text based on prior content. These two learning objectives are not in conflict and could potentially be integrated. Analogous to a child learning a new language via practicing both cloze exercises and writing assignments, the training mechanism for a model can similarly employ a multi-task strategy. Therefore, we consider enabling our model to learn both tasks concurrently. To achieve this, we adopt a unified model architecture and alternate the training objective between MLM and CLM tasks. This approach attempts to mimic the human learning process, hence helping the model acquire deeper knowledge from a limited amount of text data.

To examine the effect of integrating MLM and CLM pretraining tasks on model performance, we conducted experiments using LTG-BERT and BabyLlama\footnote{We only utilized the BabyLlama architecture and did not apply the knowledge distillation method here.} as base models, testing on the BabyLM2024 10M datasets. LTG-BERT, an Encoder-only model, and BabyLlama, a Decoder-only model, are notable architectures from the 2023 BabyLM Challenge .The results indicate that both LTG-BERT and BabyLlama showed improvements in macroaverage scores. These experiments confirm that the integration of these two pretraining objectives can positively impact model training.
\section{Related Work}
\label{sec:related-work}

\textbf{Causal Language Models} have played a pivotal role in the development of NLP, particularly in tasks involving sequence generation. The foundational work by OpenAI on the Generative Pre-trained Transformer (GPT)~\cite{gpt} marked a significant breakthrough in the use of CLMs for a variety of NLP applications. GPT~\cite{gpt} models the probability of each token in a sequence based on all preceding tokens, enabling it to perform well on tasks like text completion, machine translation, and summarization. The subsequent release of GPT-2~~\cite{gpt2} and GPT-3~\cite{gpt3} further illustrated the power of scaling CLMs. These models, with their increased parameter sizes and training data, have set new performance benchmarks in tasks like zero-shot and few-shot learning.  The GPT family firmly established the dominance of autoregressive models in generative tasks. More recently, Meta introduced the LLaMA~\cite{touvron2023llama} series, which demonstrated that highly capable CLMs could be trained efficiently on fewer parameters and less compute than earlier models like GPT-3. 
LLaMA, designed to be accessible for academic research, retains the autoregressive framework while achieving competitive performance across a range of NLP tasks.

\textbf{Masked Language Model} is a training approach used to develop deep bidirectional representations of context, often referred to as a cloze task~\cite{taylor1953cloze}. Specifically, a special token [MASK] is employed to randomly mask a proportion of input tokens, and the model is trained to predict these masked tokens. This training task was first innovatively introduced in BERT~\cite{devlin2018bert} and has been adopted in subsequent models like RoBERTa~\cite{liu2019roberta} and ALBERT~\cite{lan2019albert}. Research has also led to improvements in MLM tasks, such as in SpanBERT~\cite{joshi2020spanbert}, where the model is trained to predict spans of words instead of individual tokens, enhancing its ability to capture long-range dependencies.


\textbf{Unified modeling} refers to using a single model architecture to handle multiple training and evaluation tasks. In the T5~\cite{raffel2020exploringT5} model, various downstream tasks were reformulated as text-to-text tasks, significantly enhancing the model’s ability for multitask learning. Moreover, many related works~\cite{sanh2019hierarchicalUni1,liu2020attentionUni2} have also applied unified modeling for multitask training and evaluation, making it a common approach to improve the generalization ability of models.
UniLM~\cite{dong2019unified}, based on the BERT architecture, is one of the significant endeavors in unified modeling. By employing specific self-attention masks, UniLM controls the contextual information used during prediction. When predicting tokens, it not only trains like an autoencoding language model by leveraging the context of masked tokens but also performs left-to-right training like an autoregressive language model. Additionally, UniLM can function similarly to encoder-decoder architectures by encoding the first input text and then generating sequences from left to right. By switching the attention matrix, it seamlessly transitions between different training tasks and downstream application scenarios. 

Existing methods have unified CLM and MLM networks regarding model architecture and parameter sharing.
However, research on unifying their training objectives remains unexplored. 
This paper is the first to bridge the two classic training objectives.

\section{Methods}
\label{sec:method}

\subsection{Preliminaries}
\label{ssec:prel}

BabyLlama~\cite{timiryasov2023BabyLlamaknowledgedistillation} was proved to be effective in BabyLM2023 and is included as one of the baselines officially provided by BabyLM2024. BabyLlama~\cite{timiryasov2023BabyLlamaknowledgedistillation}  employed knowledge distillation, transferring the knowledge from two teacher models — a GPT-2 model with 705 million parameters and a LLaMA model with 360 million parameters — into a compact BabyLlama “student” model with just 58 million parameters. Given that our own replication of the BabyLlama model through distillation did not achieve ideal results, we opted to use only the BabyLlama architecture  with a parameter size of 97 million. The BabyLlama model employs the classic CLM paradigm~\cite{gpt}, where given the first $n$ tokens in a sequence, the model predicts the token at position $n+1$. The next-token prediction (NTP) training objective is to minimize the negative log-likelihood loss of predicting the next token at each timestep. To achieve this, a causal mask is applied in the self-attention mechanism. This mask is represented as a lower triangular matrix, ensuring each token can only attend to its preceding tokens. Formally, for an input sequence of length $T$, $x_1, x_2, \dots, x_T$, the corresponding attention mask $M$ is a $T \times T$ lower triangular matrix, where $M_{ij} $ indicates whether the token at position $i$ should attend to the token at position $j$. 
This masking strategy effectively prevents the model from accessing future information during training, thereby capturing the sequential order and dependencies within the data.

\begin{figure}[t]
  \centering
  \includegraphics[width=\linewidth]{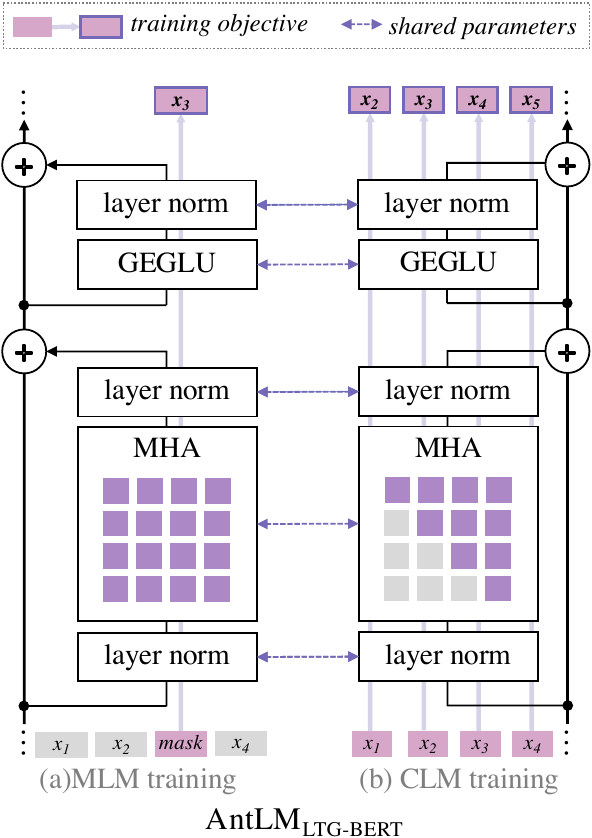}
  \caption{A diagram of AntLM$_\text{LTG-BERT}$. Based on the LTG-BERT architecture, we propose a joint MLM and CLM training objective. It is worth noting that the two objectives fully share parameters, but differ in their attention masks. The diagram also applies to AntLM$_\text{BabyLlama}$, with the difference in the architecture (e.g., positional encoding and the activation function of GLU).}
  \label{fig:main}
\end{figure}

In  BabyLM2023~\cite{conll-2023-babylm}, experiments with Boot-BERT~\cite{samuel2023meanbootbert} and ELC-BERT~\cite{charpentier2023not} demonstrated the effectiveness of the LTG-BERT~\cite{samuel2023trained} architecture. LTG-BERT is also one of the official baselines in BabyLM2024. The LTG-BERT model incorporates several key architectural improvements, including NormFormer layer normalization~\cite{shleifer2021normformer}, disentangled attention with relative position embeddings~\cite{he2020deberta}, and gated-linear activation function~\cite{shazeer2020glu}. The training objective of LTG-BERT is self-supervised Masked Language Modeling (MLM). During training, 15\% of the tokens in the input sequence are randomly selected for replacement. Of these, 80\% are masked, 10\% are substituted with random tokens, and the remaining 10\% are unchanged. The model is then trained to predict the original masked tokens based on the context. LTG-BERT explores three common masking strategies: subwords, whole words, and spans. Experimental results indicate that span-based masking yields slightly better performance compared to the other methods.

\subsection{Our Approach}
\label{ssec:method}

Inspired by the way children learn languages through both cloze exercises and writing assignments, our work constructs a unified training framework that integrates CLM and MLM. In this unified framework, we switch between the two training paradigms alternately. CLM uses a causal mask to enforce sequential dependencies and MLM employs bidirectional attention, enabling the model to predict masked tokens by leveraging both preceding and succeeding context. By combining these two training objectives, the model not only excels at autoregressive tasks like text generation but also achieves a deeper semantic understanding of language by capturing broader contextual information through bidirectional attention. 


In our approach, we integrate CLM and MLM by alternating between these training objectives during the pre-training phase. After training the model on one objective for a specified number of epochs, we switch to the other objective. The switch between training objectives is implemented by modifying the model's input and attention matrix. For the MLM task, 15\% of the tokens in the input are randomly selected and replaced. The model utilizes bidirectional attention to predict the original tokens based on the surrounding context. In contrast, for the CLM task, no token replacement is required in the input. The model employs causal attention to predict the next token based on the preceding tokens.

\section{Experiment}
\label{sec:exper}

\begin{table}[t]
	\centering
	\begin{tabular}{ccc}
		\toprule  
		\textbf{Name}&\textbf{BabyLlama}&\textbf{LTG-BERT} \\
		\midrule  
		layers&12&12 \\
        attention heads&12&12\\ 
        hidden size&768&768\\
        intermediate size&2048&2048\\
        vocabulary size&16k&16k\\
        position bucket  & -- &32 \\
		\bottomrule  
	\end{tabular}
\caption{Model Hyper-parameters.}
\label{tab:Hyper-parameter}
\end{table}
\begin{table*}[t]
    \centering
    \begin{tabular}{ccccccc}
        \toprule  
  \textbf{Model}&\textbf{Data}&\textbf{BLiMP}&\textbf{$\text{BLiMP}_\text{Supplement}$}&\textbf{EWoK}&\textbf{GLUE}&\textbf{\textit{Macro Avg.}} \\ 
        \midrule  
        BabyLlama$^\dag$& 10M & \bf 69.8 & 59.5 & 50.7 & 63.3&60.8 \\
        BabyLlama& 10M &  68.1 & 60.4 & 50.4 & 65.5  &61.1 \\
        AntLM$_\text{BabyLlama}$ & 10M&69.4&\bf 60.7 &\bf 51.1 &\bf 67.4 & \bf 62.1 \\ 
        \midrule  
        BabyLlama$^\dag$& 100M &  73.1 & 60.6 &\bf 52.1 & \bf69.0&63.7 \\
        LTG-BERT$^\dag$&100M&69.2&\bf 66.5&51.9&68.4&64.0 \\
         BabyLlama & 100M& \bf 74.9& 66.0 & 52.0 & 66.3 & \bf 64.8 \\
       
        \midrule
        LTG-BERT$^\dag$&10M&60.6&60.8&48.9&60.3&57.5 \\
        LTG-BERT&10M&62.6&\bf 65.4&62.3&64.9&63.8 \\
        AntLM$_\text{LTG-BERT}$ &10M&\bf 72.3&62.6 &  \bf 63.0 & \bf 66.0 & \bf 66.0 \\
        \bottomrule  
    \end{tabular}
    \caption{Main experimental results. The $^\dag$ indicates results from the official report. The official BabyLlama leverages knowledge distillation, while our AntLM$_\text{BabyLlama}$ is based solely on the architecture of BabyLlama without knowledge distillation methods. Due to limitations in time and resources, we have not attempted AntLM on the 100M track, this will be part of our future work. }
    \label{tab:main-result}
\end{table*}

\paragraph{Data Preprocessing}
For the data preprocessing part, we adopt the data handling procedures from the BootBERT~\cite{samuel2023meanbootbert} method, which performed well in the previous round of BabyLM Challenge. Preprocessing includes steps like normalizing punctuation, reconstructing sentence structures, and removing duplicate text. These preprocessing steps help ensure cleaner and more structured input data, contributing to better model performance.

\paragraph{Baselines}
We adopt the official baseline provided by the BabyLM Challenge as our benchmark, using the results achieved by the best-performing models from the previous round, namely LTG-BERT and BabyLlama, see Table~\ref{tab:main-result}.

\paragraph{Experiment Settings}
In our experiments, we used both the BabyLlama and LGT-BERT models to evaluate the performance of a hybrid training strategy combining Causal Language Modeling (CLM) and Masked Language Modeling (MLM). For both model architectures, we used the same set of parameters, as shown in the table~\ref{tab:Hyper-parameter} and optimized the training process using the AdamW optimizer. Additionally, we employed the bfloat16 data type to enhance computational efficiency. For the BabyLlama model, we used a batch size of 512 with an initial learning rate set to $7 \times 10^{-4}$.  The learning rate scheduler followed a cosine decay during the CLM training phase and a cosine with restarts scheduler during the MLM phase, with the number of cycles set to every four epochs . For the LGT-BERT model, we employed a batch size of 1024, with an initial learning rate of $5 \times 10^{-4}$. In all training phases, we used a cosine with restarts scheduler, with the num cycles set to 4. Our hyperparameters were determined through multiple experiments, building upon the hyperparameter settings from the previous works~\cite{timiryasov2023BabyLlamaknowledgedistillation,samuel2023trained} to find the optimal values.
The training process alternated between CLM and MLM objectives over multiple epochs. We used the notation ``$x$\_CLM + $y$\_MLM..." to indicate that, \textit{in sequential order}, $x$ epochs are trained in the CLM training mode, followed by $y$ epochs in the MLM training mode, and so on.

\definecolor{CLM}{RGB}{173, 136, 198}
\definecolor{MLM}{RGB}{255, 230, 230}
\begin{table*}[t]
    \centering
    \begin{tabular}{ccccc}
        \toprule  
        \textbf{Training Stage}&\textbf{BLiMP}&\textbf{$\text{BLiMP}_\text{Supplement}$}&\textbf{EWoK}&\textbf{Avg.} \\ 
        \midrule  

        \rowcolor{gray!0}  
        \multicolumn{5}{c}{\textbf{AntLM$_\text{BabyLlama}$}} \\
        \makecell[l]{
        \begin{tikzpicture}[baseline=(current bounding box.center)]
             \fill[fill=CLM] (0, 0) rectangle (2.4, 0.4);
             \draw[opacity=0] (2.4, 0) rectangle (7.2, 0.4);
              \node at (1.2, 0.2) {8};
        \end{tikzpicture}
        }
        & 68.2& 56.7& 50.5 & 58.5\\

        \makecell[l]{
        \begin{tikzpicture}[baseline=(current bounding box.center)]
             \fill[fill=MLM] (0, 0) rectangle (4.8, 0.4);
             \draw[opacity=0] (4.8, 0) rectangle (7.2, 0.4);
             \node at (2.4, 0.2) {16};
         \end{tikzpicture} }
        & 56.8&58.4&57.2 & 57.5\\ 
         
        \makecell[l]{
        \begin{tikzpicture}[baseline=(current bounding box.center)]
             \fill[fill=CLM] (0, 0) rectangle (7.2, 0.4);
             \node at (3.6, 0.2) {24};
         \end{tikzpicture}}
         &68.1&60.4&50.4&59.6\\

         \makecell[l]{
         \begin{tikzpicture}[baseline=(current bounding box.center)]
             \fill[fill=MLM] (0, 0) rectangle (7.2, 0.4);
             \node at (3.6, 0.2) {24};
         \end{tikzpicture}  }
         & 56.9& 57.8&\textbf{58.3} & 57.7\\

        \makecell[l]{
        \begin{tikzpicture}[baseline=(current bounding box.center)]
             \fill[fill=CLM] (0, 0) rectangle (1.2,0.4);
             \node at (0.6, 0.2) {4};
             \fill[fill=MLM] (1.2, 0) rectangle (6, 0.4);
             \node at (3.6, 0.2) {16};    
             \fill[fill=CLM] (6, 0) rectangle (7.2, 0.4);
             \node at (6.6, 0.2) {4};
         \end{tikzpicture} }
         &\textbf{69.4}&\textbf{60.7}&51.1 & \textbf{60.4}\\

        \midrule  

        \rowcolor{gray!0}  
        \multicolumn{5}{c}{\textbf{AntLM$_\text{LTG-BERT}$}} \\

        \makecell[l]{
        \begin{tikzpicture}[baseline=(current bounding box.center)]
             \fill[fill=CLM] (0, 0) rectangle (1.2, 0.4);
             \draw[opacity=0] (1.2, 0) rectangle (7.2, 0.4);
             \node at (0.6, 0.2) {12};
         \end{tikzpicture}}
         & 69.9 & 56.4 & 50.8 & 59.0\\

        \makecell[l]{
        \begin{tikzpicture}[baseline=(current bounding box.center)]
             \fill[fill=MLM] (0, 0) rectangle (6.0, 0.4);
             \draw[opacity=0] (6.0, 0) rectangle (7.2, 0.4);
             \node at (3.0, 0.2) {60};
         \end{tikzpicture}}
         & 62.8 & \textbf{63.5} & 64.2 & 63.5\\

        \makecell[l]{
        \begin{tikzpicture}[baseline=(current bounding box.center)]
             \fill[fill=CLM] (0, 0) rectangle (7.2, 0.4);
             \node at (3.6, 0.2) {72};
         \end{tikzpicture}}
         & 70.0 & 57.2 & 51.9 & 57.9\\

         \makecell[l]{
         \begin{tikzpicture}[baseline=(current bounding box.center)]
             \fill[fill=MLM] (0, 0) rectangle (7.2, 0.4);
             \node at (3.6, 0.2) {72};
         \end{tikzpicture}  }
         & 69.4 & 61.1 & \textbf{64.5} & 65.0\\

        \makecell[l]{
        \begin{tikzpicture}[baseline=(current bounding box.center)]
             \fill[fill=CLM] (0, 0) rectangle (0.6,0.4);
             \node at (0.3, 0.2) {6};
             \fill[fill=MLM] (0.6, 0) rectangle (6.6, 0.4);
             \node at (3.6, 0.2) {60};    
             \fill[fill=CLM] (6.6, 0) rectangle (7.2, 0.4);
             \node at (6.9, 0.2) {6};
         \end{tikzpicture}  }
         & \textbf{72.3} & 62.5 & 63.0 & \textbf{66.0}\\

        \bottomrule  
    \end{tabular}
    \caption{The effect of integrating \colorbox{CLM}{CLM} and \colorbox{MLM}{MLM} training objectives on BabyLlama and LTG-BERT.}
    \label{tab:main-experiment}
\end{table*}
\subsection{Main Results}
\begin{figure*}[ht]
    \centering
    \begin{subfigure}{0.49\textwidth} 
        \includegraphics[width=\linewidth]{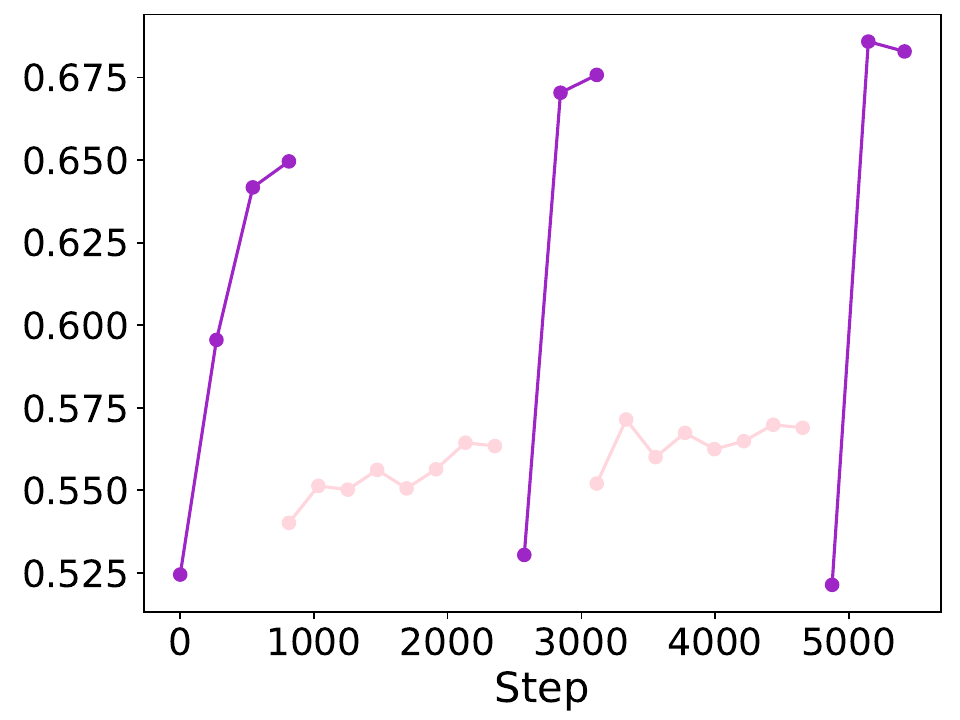}
      \caption{BLiMP}
        \label{fig:sub1}
    \end{subfigure}
    \begin{subfigure}{0.49\textwidth}
        \includegraphics[width=\linewidth]{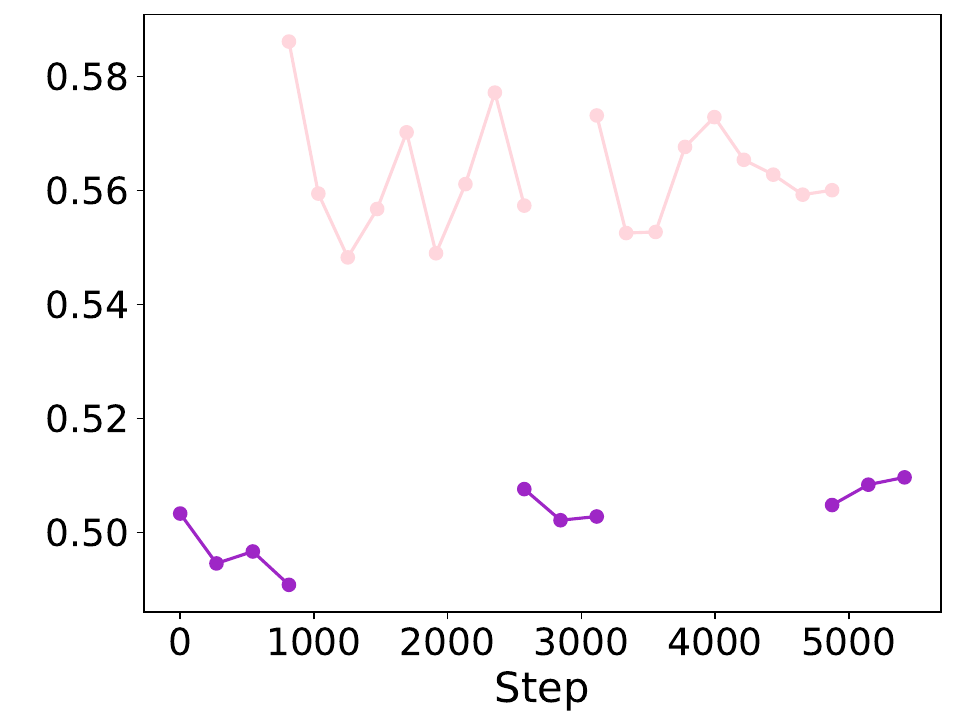}
        \caption{EWok}
        \label{fig:sub3}
    \end{subfigure}
    \caption{The phased experimental results on three datasets. The evaluation line chart for each stage of ``3\_\colorbox{CLM}{CLM} + 8\_\colorbox{MLM}{MLM} + 2\_\colorbox{CLM}{CLM} + 8\_\colorbox{MLM}{MLM} + 3\_\colorbox{CLM}{CLM}'' on the BabyLlama model. The reason for the discontinuity in evaluation results between training phases is that we applied the evaluation method corresponding to the specific task categories at each stage of the training process.}
    \label{fig:example}
\end{figure*}

In this section, we evaluate the performance of BabyLlama and LTG-BERT across multiple benchmarks, including BLiMP, BLiMP Supplement, EWoK, and GLUE. Our experiments primarily focus on assessing the impact of integrating CLM and MLM training objectives on the overall results, comparing the baseline performance of both BabyLlama and LTG-BERT with the configurations we propose. 

As shown in Table \ref{tab:main-result}, our models with integrated training objectives consistently outperform the official baseline scores on both the LTG-BERT and BabyLlama models. Notably, the improvements on LTG-BERT are particularly significant, demonstrating the effectiveness of our approach. 
To further validate the effectiveness of alternating training objectives CLM and MLM, we conducted an in-depth experiment with the BabyLlama model. Given the lengthy training times associated with the GLUE dataset, we opted to evaluate our results on the BLiMP, BLiMP Supplement, and EWoK datasets. As shown in Table \ref{tab:main-experiment}, the model trained with the \textit{4\_CLM+16\_MLM+4\_CLM} strategy significantly outperformed those trained solely with \textit{8\_CLM} or \textit{16\_MLM}. This finding indicates that combining these two training objectives enables the model to simultaneously acquire bidirectional context understanding and sequence generation capabilities. Under the same training epochs, the \textit{4\_CLM+16\_CLM+4\_CLM} combination demonstrated clear advantages over the pure \textit{24\_CLM} and \textit{24\_MLM} models, further confirming that the integration of these two training objectives is crucial for achieving optimal performance, highlighting the complementary relationship between CLM and MLM. We also conducted similar experiments on the LTG-BERT, the results are shown on same Table.

\begin{table*}[htb]
    \centering
    \begin{tabular}{ccccc}
        \toprule  
        \textbf{Training Stage}&\textbf{BLiMP}&\textbf{$\text{BLiMP}_\text{Supplement}$}&\textbf{EWoK} &\textbf{Avg.} \\ 
        \midrule  
        \rowcolor{gray!0}  
        \multicolumn{5}{c}{\textbf{AntLM$_\text{BabyLlama}$}} \\
        \makecell[l]{
        \begin{tikzpicture}[baseline=(current bounding box.center)]
             \fill[fill=CLM] (0, 0) rectangle (2.4, 0.4); 
             \fill[fill=MLM] (2.4, 0) rectangle (7.2, 0.4);
             \node at (1.2, 0.2) {8};
             \node at (4.8, 0.2) {16};
         \end{tikzpicture}
        }& 68.2& 56.7& 50.5 & 58.5\\
        
        \makecell[l]{
        \begin{tikzpicture}[baseline=(current bounding box.center)]
             \fill[fill=MLM] (0, 0) rectangle (4.8, 0.4); 
             \fill[fill=CLM] (4.8, 0) rectangle (7.2, 0.4);
             \node at (2.4, 0.2) {16};
             \node at (6.0, 0.2) {8};
         \end{tikzpicture}
        }& 68.4&\textbf{61.1}&50.1 & 59.9\\ 
        \makecell[l]{
        \begin{tikzpicture}[baseline=(current bounding box.center)]
             \fill[fill=CLM] (0, 0) rectangle (1.2, 0.4); 
             \fill[fill=MLM] (1.2, 0) rectangle (6.0, 0.4);
             \fill[fill=CLM] (6.0, 0) rectangle (7.2, 0.4);
             \node at (0.6, 0.2) {4};
             \node at (3.6, 0.2) {16};
             \node at (6.6, 0.2) {4};
         \end{tikzpicture}
        }&\textbf{69.4}&60.7&\textbf{51.1} & \textbf{60.4}\\

        \makecell[l]{
        \begin{tikzpicture}[baseline=(current bounding box.center)]
             \fill[fill=MLM] (0, 0) rectangle (2.4, 0.4);
             \fill[fill=CLM] (2.4, 0) rectangle (4.8, 0.4);
             \fill[fill=MLM] (4.8, 0) rectangle (7.2, 0.4);
             \node at (1.2, 0.2) {8};
             \node at (3.6, 0.2) {8};
             \node at (6, 0.2) {8};
         \end{tikzpicture}
        }
        &67.2&59.2&50.2& 58.9\\

        \makecell[l]{
        \begin{tikzpicture}[baseline=(current bounding box.center)]
             \fill[fill=CLM] (0, 0) rectangle (1.2, 0.4); 
             \fill[fill=MLM] (1.2, 0) rectangle (3.6, 0.4);
             \fill[fill=CLM] (3.6, 0) rectangle (4.8, 0.4);
             \fill[fill=MLM] (4.8, 0) rectangle (7.2, 0.4);
             \node at (0.6, 0.2) {4};
             \node at (2.4, 0.2) {8};
             \node at (4.2, 0.2) {4};
             \node at (6, 0.2) {8};
         \end{tikzpicture}
        }
        &68.8&60.6&50.7& 60.0\\

        \makecell[l]{
        \begin{tikzpicture}[baseline=(current bounding box.center)]
             \fill[fill=MLM] (0, 0) rectangle (2.4, 0.4); 
             \fill[fill=CLM] (2.4, 0) rectangle (3.6, 0.4);
             \fill[fill=MLM] (3.6, 0) rectangle (6, 0.4);
             \fill[fill=CLM] (6, 0) rectangle (7.2, 0.4);
             \node at (1.2, 0.2) {8};
             \node at (3, 0.2) {4};
             \node at (4.8, 0.2) {8};
             \node at (6.6, 0.2) {4};
         \end{tikzpicture}
        }
        &68.6&59.1&51.0& 59.6\\
        
        \makecell[l]{
        \begin{tikzpicture}[baseline=(current bounding box.center)]
             \fill[fill=CLM] (0, 0) rectangle (0.9, 0.4); 
             \fill[fill=MLM] (0.9, 0) rectangle (3.3, 0.4);
             \fill[fill=CLM] (3.3, 0) rectangle (3.9, 0.4);
             \fill[fill=MLM] (3.9, 0) rectangle (6.3, 0.4);
             \fill[fill=CLM] (6.3, 0) rectangle (7.2, 0.4);
             \node at (0.45, 0.2) {3};
             \node at (2.1, 0.2) {8};
             \node at (3.6, 0.2) {2};
             \node at (5.1, 0.2) {8};
             \node at (6.75, 0.2) {3};
         \end{tikzpicture}
        }
        &69.3&60.1&50.8& 60.1\\

        \makecell[l]{
        \begin{tikzpicture}[baseline=(current bounding box.center)]
             \fill[fill=CLM] (0, 0) rectangle (0.3, 0.4); 
             \fill[fill=MLM] (0.3, 0) rectangle (1.2, 0.4);
             \fill[fill=CLM] (1.2, 0) rectangle (1.5, 0.4);
             \fill[fill=MLM] (1.5, 0) rectangle (2.4, 0.4);
             \fill[fill=CLM] (2.4, 0) rectangle (2.7, 0.4);
             \fill[fill=MLM] (2.7, 0) rectangle (3.3, 0.4);
             \fill[fill=CLM] (3.3, 0) rectangle (3.6, 0.4);
             \fill[fill=MLM] (3.6, 0) rectangle (4.2, 0.4);
             \fill[fill=CLM] (4.2, 0) rectangle (4.5, 0.4);
             \fill[fill=MLM] (4.5, 0) rectangle (5.1, 0.4);
             \fill[fill=CLM] (5.1, 0) rectangle (5.4, 0.4);
             \fill[fill=MLM] (5.4, 0) rectangle (6.0, 0.4);
             \fill[fill=CLM] (6.0, 0) rectangle (6.3, 0.4);
             \fill[fill=MLM] (6.3, 0) rectangle (6.9, 0.4);
             \fill[fill=CLM] (6.9, 0) rectangle (7.2, 0.4);
             \node at (0.15, 0.2) {1};
             \node at (0.75, 0.2) {3};
             \node at (1.35, 0.2) {1};
             \node at (1.95, 0.2) {3};
             \node at (2.55, 0.2) {1};
             \node at (3, 0.2) {2};
             \node at (3.45, 0.2) {1};
             \node at (3.9, 0.2) {2};
             \node at (4.35, 0.2) {1};
             \node at (4.8, 0.2) {2};
             \node at (5.25, 0.2) {1};
             \node at (5.7, 0.2) {2};
             \node at (6.15, 0.2) {1};
             \node at (6.60, 0.2) {2};
             \node at (7.05, 0.2) {1};
         \end{tikzpicture}
        }& 67.3& 55.2&50.4& 57.6\\
        
        \bottomrule  
        
    \end{tabular}
    \caption{The effect of alternating frequency (low or high) and alternating order of \colorbox{CLM}{CLM} and \colorbox{MLM}{MLM} training objectives on BabyLlama. All were trained for a total of 24 epochs.}
    \label{tab:ab_bl}
\end{table*}
 
Additionally, we explored the performance of these training modes across different datasets. As shown in Figure~\ref{fig:example}, MLM performs significantly better on the EWoK dataset, while CLM exhibits more pronounced and sensitive results on the BLiMP dataset. This indicates that different training approaches have varying impacts on distinct datasets. Thus, the integrated experiments that combine both training methods can better leverage their strengths and enhance overall performance.

\subsection{Ablation Study}

To investigate the effects of various factors on the evaluation task results within the integrated experiments, we conducted ablation studies focusing on two variables: alternating frequency and alternating order. In the BabyLlama model, we maintained a constant total number of training epochs at 24 (8 epochs for the CLM phase and 16 epochs for the MLM phase). Specifically, for the alternating order, we adjusted the alternating sequence of training between the CLM and MLM phases while keeping the overall epoch count unchanged. For alternating frequency, we divided the training process into more frequent alternating stages. 
The experimental results, as shown in Table \ref{tab:ab_bl}, indicate that variations in these two factors do not lead to significant declines in evaluation outcomes, suggesting that our approach is stable. We hypothesize that the decrease in performance with an increased frequency of alternations may be attributed to smaller epoch sizes in each training phase, which could hinder convergence on the respective tasks.

Furthermore, we found that the best performance was achieved when the CLM training phase was placed at both the beginning and the end of the training sequence, which could be due to the greater impact of CLM compared to MLM. Although CLM does not inherently have a higher performance ceiling (as last year’s winner was an MLM-based model), but it converges more rapidly. CLM performs sequential prediction training on every token, while MLM focuses only on masked tokens. Thus, we suggest that CLM captures more learning within a single epoch than MLM.


\section{Conclusion}

In this study, we propose AntLM, a model that applies to multiple natural language-related tasks in the BabyLM Challenge by alternating between Causal Language Modeling (CLM) and Masked Language Modeling (MLM) during training. Experimental results demonstrate that AntLM achieves either superior or comparable performance to the baseline across all evaluation tasks. 

Additionally, we found that CLM and MLM have different impacts on various evaluation tasks, suggesting that these training tasks guide the model to learn distinct aspects of human language. We believe this difference is the key reason why integrated training yields effective results, as the model benefits from the knowledge learned from both training approaches. This finding also raises an intriguing question: do different training tasks allow models to capture only specific portions of natural language knowledge? Due to resource limitations, we were unable to explore additional ideas and approaches in this study. In future work, we plan to address these limitations by expanding our resources and support, allowing us to further investigate these potential directions.

Moreover, we conducted experiments with varying numbers and sequences of alternating training, and the results suggest that specific integrated training methods are more effective in achieving optimal evaluation outcomes.

\bibliography{custom}

\appendix


\end{document}